\documentclass[runningheads]{llncs}

\usepackage{eccv}

\usepackage{eccvabbrv}

\usepackage{graphicx}
\usepackage{booktabs}

\usepackage[accsupp]{axessibility}  

\usepackage{hyperref}

\usepackage{orcidlink}

\definecolor{mygreen}{RGB}{0, 170, 0} 

\usepackage{bbm} 
\usepackage{marvosym}

\usepackage{algorithm}

\usepackage{xcolor} 
\usepackage{pifont} 
\usepackage{verbatim} 
\usepackage{multirow}
\usepackage{nameref} 
\usepackage{kotex} 
\usepackage{colortbl}
\usepackage{amsmath} 
\usepackage{amssymb} 

\usepackage{subcaption} 
\usepackage{afterpage}

\usepackage{rotating}

\usepackage{caption} 

\usepackage[absolute,overlay]{textpos} 
\usepackage{calc}

\usepackage{algpseudocode}

\begin{document}

\title{Rethinking LiDAR Domain Generalization: Single Source as Multiple Density Domains}

\titlerunning{Rethinking LiDAR Domain Generalization}

\author{Jaeyeul Kim\inst{*}\orcidlink{0000-0002-7765-4972} \and
Jungwan Woo\inst{*}\orcidlink{0000-0001-6099-5490} \and
Jeonghoon Kim\orcidlink{0000-0002-7568-4115} \and
Sunghoon Im\textsuperscript{(\Letter)}\orcidlink{0000-0001-9776-8101}}

\authorrunning{J. Kim et al.}

\institute{DGIST, Daegu, South Korea\\
\email{\{jykim94, friendship1, jeonghoon, sunghoonim\}@dgist.ac.kr}
}

\renewcommand*{\thefootnote}{*}
\footnotetext[1]{J. Kim and J. Woo---Authors contributed equally to this work.
}

\maketitle

\begin{abstract}
In the realm of LiDAR-based perception, significant strides have been made, yet domain generalization remains a substantial challenge. The performance often deteriorates when models are applied to unfamiliar datasets with different LiDAR sensors or deployed in new environments, primarily due to variations in point cloud density distributions.
To tackle this challenge, we propose a Density Discriminative Feature Embedding (DDFE) module, capitalizing on the observation that a single source LiDAR point cloud encompasses a spectrum of densities.
The DDFE module is meticulously designed to extract density-specific features within a single source domain, facilitating the recognition of objects sharing similar density characteristics across different LiDAR sensors.
In addition, we introduce a simple yet effective density augmentation technique aimed at expanding the spectrum of density in source data, thereby enhancing the capabilities of the DDFE.
Our DDFE stands out as a versatile and lightweight domain generalization module.
It can be seamlessly integrated into various 3D backbone networks, where it has demonstrated superior performance over current state-of-the-art domain generalization methods.
Code is available at \url{https://github.com/dgist-cvlab/MultiDensityDG}.
\end{abstract}

\begin{figure}[t]
    \centering
    \includegraphics[width=0.99\textwidth]
    {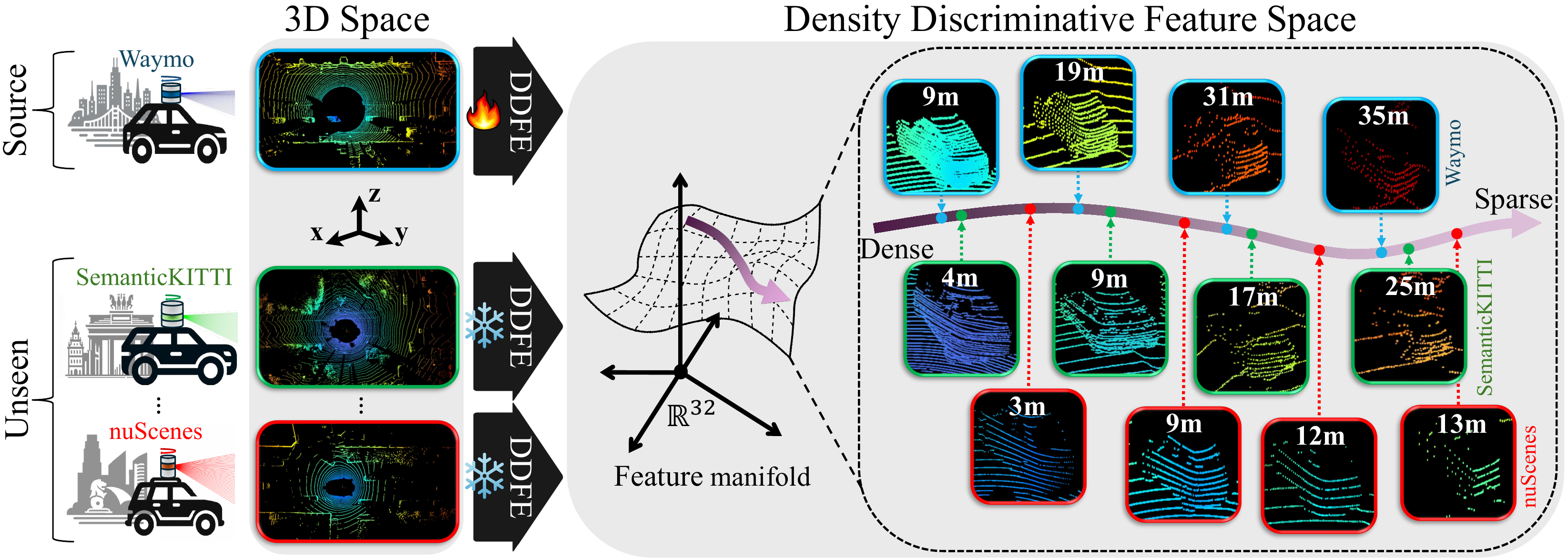}
    \caption{\textbf{Motivation of our DDFE: Leveraging Diverse Densities in a Source LiDAR for Domain Generalization} - Despite the apparent differences in density distributions between Waymo (64-channel), SemanticKITTI (64-channel), and nuScenes (32-channel), they share regions of overlapping density distributions.
    For example, the observation that a vehicle at 35 meters in Waymo (top) has a similar density to one at 25 meters in SemanticKITTI (middle) and 12 meters in nuScenes (bottom) underscores this phenomenon.
    This understanding of varying local density within a source domain serves as a foundation of our domain generalization method.
    The proposed DDFE transforms features from 3D space to a unified density space without additional training on unseen data, enhancing domain generalization performance.
    }
    \label{fig_teaser}
\end{figure}

\section{Introduction}
\label{sec:intro}

Light Detection and Ranging (LiDAR) provides detailed 3D data, making it indispensable for environmental perception in autonomous vehicles. 
Among the various LiDAR-based perception tasks, semantic segmentation plays a crucial role in understanding the driving scene by classifying each point into multiple classes. 
While LiDAR-based semantic segmentation~\cite{Zhu2021Cylindrical, Hou2022PVKD, Ando2023Rangevit} have been widely studied, their impressive performance is often constrained to scenarios where source and target datasets align perfectly.
However, mismatches between these datasets can lead to significant performance declines. These are mainly attributed to two primary factors: environmental variations as exemplified by the Waymo~\cite{Sun2020waymo} in the USA, the SemanticKITTI~\cite{Behley2019SemanticKITTI} in Germany, and the nuScenes~\cite{Caesar2020NUSCENES} in Singapore; and sensor-induced discrepancies, including differences in the number of beams and the field of view (FOV).

Numerous Unsupervised Domain Adaptation (UDA) studies~\cite{Saltori2022COSMIX, yuan2022category, kong2023conda} address these issues but require additional fine-tuning whenever the target domain changes.
In contrast, models deployed with robust Domain Generalization (DG) techniques are increasingly sought after for their potential to enable real-time autonomous driving systems without needing constant fine-tuning.
Nevertheless, this crucial field remains under-researched, and its potential has yet to be fully realized.
Existing study~\cite{Kim2023SDDG} pinpoints the point cloud density distribution as a prime performance hindrance. 
While some solutions like point cloud sampling and completion-based methods attempt to remedy this, they fall short in achieving the desired performance~\cite{Yi2021Complete} and
require sequentially labeled data and knowledge of ego-motion~\cite{Ryu2023LiDomAug}.

In this paper, we introduce a novel perspective in domain generalization by addressing the challenges posed by density variations across different LiDAR sensors. 
Previous studies~\cite{Kim2023SDDG, wei2022beamdrop, hu2023density} have primarily focused on global density differences, typically considering datasets from 64-channel LiDARs like Waymo~\cite{Sun2020waymo} denser than those from 32-channel LiDARs like nuScenes~\cite{Caesar2020NUSCENES}. 
However, these approaches oversimplify the inherent complexity of LiDAR data, which exhibits a wide range of density spectra over distance, and interpret them as a single global density value.
Contrary to the simplified view of global density comparisons, our research recognizes that LiDAR point clouds are composed of regions with varying densities, as shown in~\cref{fig_teaser}. These variations are influenced by the distance of objects from the LiDAR sensor, leading to a more complex and nuanced understanding of density within LiDAR data.

To do so, we propose a Density Discriminative Feature Embedding (DDFE) module, designed to enhance the domain generalization capabilities of LiDAR-based segmentation networks by exploiting these density variations.
The DDFE module incorporates a beam density estimation module that encodes the specific densities for each 3D voxel, enabling refined density discrimination across regions.
The features are then modulated using attention mechanisms guided by beam density, further enhancing the model's ability to generalize across domains with varying density characteristics.
Furthermore, we introduce a density soft clipping technique.
This method constrains the density spectrum, ensuring it does not encompass density distributions from unseen domains that are absent in source datasets.
To supplement this, we use density augmentation to widen the density spectrum of the source data, thereby enhancing its domain generalization capabilities.
Extensive experiments validate the superiority of our method over the conventional domain adaptation and generalization methods across multiple 3D backbone networks.

In summary, our primary contributions include:
\begin{itemize}
\item We introduce a new perspective for domain generalization to overcome density variations caused by different LiDAR sensors, utilizing the diverse densities within point clouds observed in the source domain.
\item We propose a density discriminative feature embedding module designed to identify areas with similar density distribution and amplify relevant features.
\item We present a simple yet effective data augmentation strategy, aimed at broadening the density spectrum of source data.
\item Extensive experiments demonstrate that our method outperforms state-of-the-art Domain Generalization and Domain Adaptation methods.
\end{itemize}

\section{Related Work}

\subsection{LiDAR-based Semantic Segmentation} 
\label{sec:related1}
LiDAR point clouds pose unique challenges due to their irregular, unordered, and unstructured nature.
Consequently, approaches to represent point cloud data have been broadly classified into three categories: Projection-based, point-based, and voxel-based methods.
Projection-based methods transform a 3D point cloud into a 2D representation through either a spherical or bird's-eye view projection.
By doing so, they can utilize lightweight models like 2D convolution~\cite{Cortinhal2020SalsaNext, Zhang2020Polarnet, MASS2022, RangeSeg2023} or transformer~\cite{Ando2023Rangevit, TransRVNet2023} rather than 3D convolution.
Nevertheless, such methods face the inherent limitation of 2D kernels not preserving the 3D geometric information of the real world.
Point-based methods~\cite{Thomas2019KPConv, Hu2020RandLA, fan2021scf} directly extract features from the point cloud. These methods utilize all the 3D spatial information without distortion.
However, the trade-off is a substantial demand on memory and computational resources.
In voxel-based approaches~\cite{graham20183d, Choyi2019MINKOWSKI, Zhu2021Cylindrical, Hou2022PVKD}, point clouds are quantized into 3D grids, known as voxels, and features are extracted using 3D convolutional methods.
Alongside sparse convolution~\cite{graham2017submanifold}, voxel-based methods achieve high performance by maintaining an actual geometric receptive field while ensuring an efficient computational load.

\begin{table}[t]
\caption{LiDAR configuration for each dataset 
($f_{\text{min}}$, $f_{\text{max}}$: the minimum and maximum LiDAR field of view,  $H_{\text{b}}$, $V_{\text{b}}$: the number of horizontal and vertical beams).
}
\centering
\resizebox{0.9\textwidth}{!}{ 
\begin{tabular}{c|c|c|c|c|c} 
                   & Waymo~\cite{Sun2020waymo}            & SemanticKITTI~\cite{Behley2019SemanticKITTI}   & nuScenes~\cite{Caesar2020NUSCENES}         & Pandaset~\cite{Xiao2021Pandaset}         & SemanticPOSS~\cite{Pan2020SemanticPOSS}     \\ \hline
$H_{\text{b}}$ & 2560             & 2048            & 1080             & 1800              & 1800              \\
$V_{\text{b}}$ & 64               & 64              & 32               & 64              & 40              \\
$\left[f_{\text{min}}, f_{\text{max}}\right](^\circ)$ & $\left[-17.6, +2.4\right]$ & $\left[-24.8, +2.0\right]$ & $\left[-30.0, +10.0\right]$ & $\left[-25.0, +15.0\right]$ & $\left[-16.0, +7.0\right]$ \\
\end{tabular}
}
\label{tab:config}
\end{table}

\subsection{LiDAR-based Domain Adaptation}
Variations in LiDAR sensor configurations, as shown in \cref{tab:config}, or alterations in operating environments can lead to differences in the distribution of point cloud data, which, in turn, may adversely affect the perception capabilities of autonomous driving systems.
To tackle this challenge without constructing new training datasets for every sensor type and location, research has increasingly focused on unsupervised domain adaptation techniques~\cite{Wu2019Squeezesegv2, Saltori2020SFUDA, yang2021st3d, Saltori2022COSMIX, yuan2022category, kong2023conda, peng2023cl3d}. These methods strive to preserve performance in the target domain by utilizing labeled data from a source domain alongside unlabeled data from the target domain.
C\&L~\cite{Yi2021Complete} leverages sequential data to train a voxel completion network and segments on the reconstructed canonical domain to overcome domain discrepancies.
Rochan \textit{et al.}~\cite{Rochan2022SSGA} propose a range-view-based unsupervised domain adaptation method, aligning beam positions between training and target data.
LiDAR-UDA~\cite{shaban2023lidaruda} employs beam subsampling to mimic different LiDAR sensors, and utilizes cross-frame ensembling and a Learned Aggregation Model to acquire improved pseudo labels.
Although these DA methods demonstrate effectiveness, they share a common limitation: the need for individual fine-tuning with target data upon each domain shift.

\subsection{LiDAR-based Domain Generalization} \quad
Domain Generalization~\cite{Lehner2022VFIELD, Xiao2023WILDDG, Kim2023SDDG, Ryu2023LiDomAug, sanchez2023domain, li2023bev, saltori2023walking} aims to improve the performance in scenarios where the target domain data, unseen during training, is encountered.
DGLSS~\cite{Kim2023SDDG} leverages sparsity invariance feature consistency and bridges the semantic correlation consistency between the source data and the sparse domain generated by beam sampling. 
LiDomAug~\cite{Ryu2023LiDomAug} aggregates multiple frames to generate dense world models and augment data by sampling through randomized LiDAR configurations with additional knowledge of ego-motion and the sequentially labeled data.
BEV-DG~\cite{li2023bev} employs a bird’s-eye view for enhanced cross-modal learning and develops density-maintained vector modeling for efficient learning of domain-invariant features.
LiDOG \cite{saltori2023walking} utilizes semantic priors in a 2D bird’s-eye view to extract domain-agnostic features.
Sanchez \textit{et al.}~\cite{sanchez2023domain} introduce a label propagation method that integrates multi-frame aggregation and ego-motion, although it demands significant computational resources due to the adaptation of KPConv~\cite{Thomas2019KPConv}.
Distinctively, our approach advances a single-frame domain generalization approach that considers the inherent characteristics of LiDAR. Consequently, our method offers the distinct advantage of bypassing the need for ego-motion and sequentially labeled data during the training and inference phases.

\section{Method}
\begin{figure*}[t] 
\begin{center}
\includegraphics[width=0.97\textwidth]{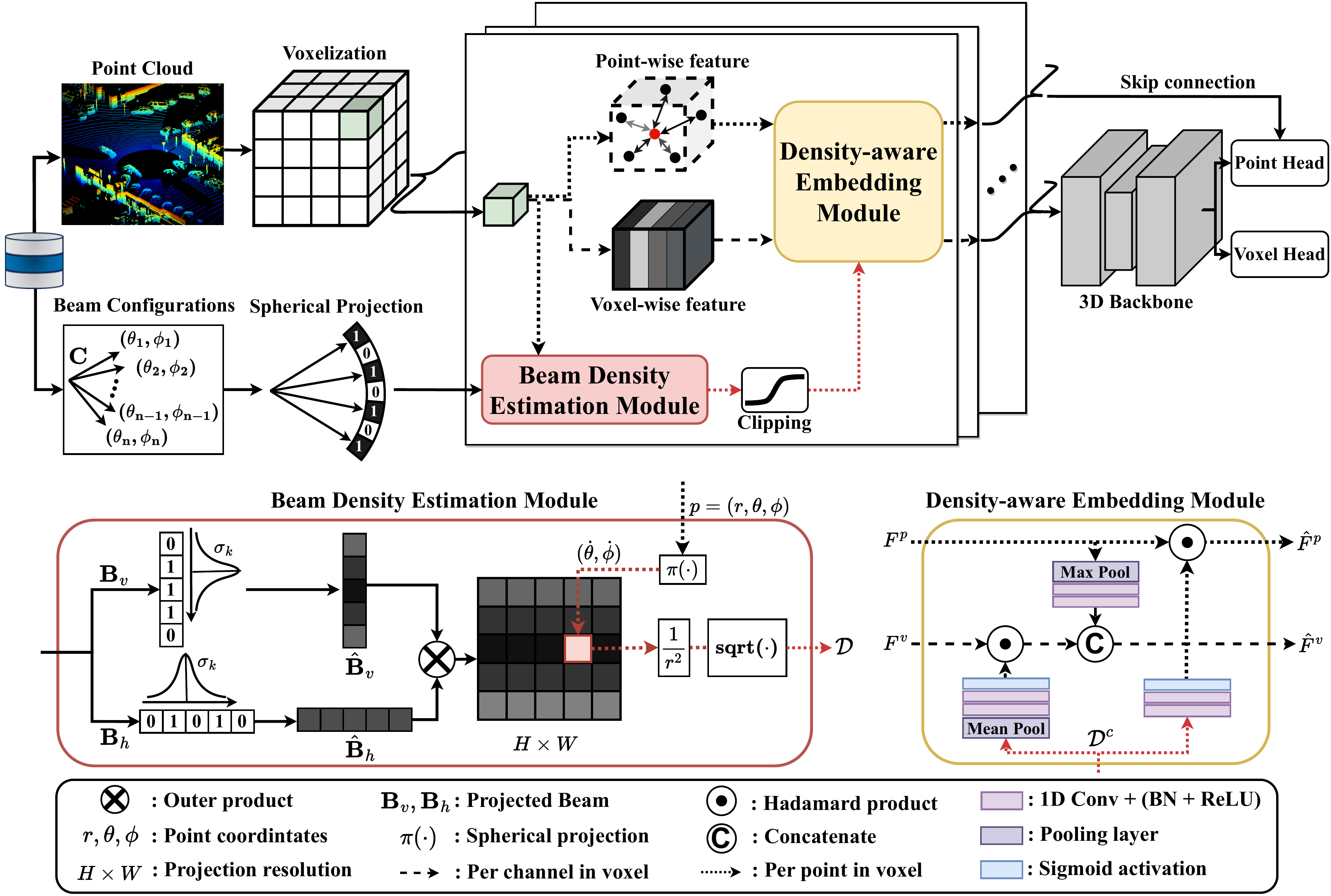}
\end{center}
\caption{Overview of the DDFE pipeline.}
\label{figure_overview}
\end{figure*}

In this section, we detail our domain generalization method for LiDAR semantic segmentation, which capitalizes on density variations observable within a single source LiDAR.
The proposed Density Discriminative Feature Embedding (DDFE) module is primarily composed of four components: Point-voxel feature encoding as detailed in~\cref{sec:PV_Encoding}, Beam density estimation module in~\cref{sec:BeamEst}, Density soft clipping in~\cref{sec:Density clipping}, and Density-aware embedding module in~\cref{sec:DensityEmb}.
Additionally, \cref{sec:data_augmentation} outlines our density augmentation technique.
The overall framework of the proposed DDFE module is shown in~\cref{figure_overview}.

\subsection{Point-voxel feature encoding}
\label{sec:PV_Encoding}
We introduce a technique to extract generalized representations across domains from point clouds.
The intensity distribution in point clouds is influenced by the specific LiDAR sensor used, which can adversely affect the performance of perception models during sensor transitions due to the variance in intensity information. 
Thus, we omit the use of LiDAR intensity values to enhance domain generalization performance. 
Furthermore, each LiDAR sensor has a unique distribution of measurement errors, particularly within the intra-voxel region.
To minimize variance induced by localized sensing noise, we incorporate a direct voxel-wise feature encoding into the PointNet-based feature extraction~\cite{qi2017pointnet}.
This direct encoding, which bypasses localized information within the voxels, effectively dismisses grid size variations, such as those of 20cm.
Given a point cloud $\mathbf{P} = \{p_i \in \mathbb{R}^{3}~|~i=1,...,N\}$ containing $N$ 3D points, we initially partition it into $M$ 3D voxels. For each voxel $v_j$, it is defined as a set of points and is represented by $\mathbf{V} = \{v_j \in \mathbb{R}^{L\times 3}~|~j = 1,...,M\}$, where $L$ indicates the number of points contained in each voxel.
The coordinates of these points are subsequently transformed into the form of $\left(\cos(\theta_j), \sin(\theta_j), \phi_j, r_j\right)$, where $(\theta_j, \phi_j, r_j)$ correspond to spherical coordinates. These transformed coordinates are then encoded into voxel-wise features $F^{v} \in \mathbb{R}^{M\times 16}$ through an MLP.
We also encode point-wise features $F^{p} \in \mathbb{R}^{N\times 16}$, which are essential to produce point-wise outputs.
For each point $k$ in voxel $v_j$ centered at $(x^c_j, y^c_j, z^c_j)$, we compute the offset as $(x_{jk} - x^c_j, y_{jk} - y^c_j, z_{jk} - z^c_j)$, with $k=1,...,L_{j}$. Here, $L_{j}$ represents the number of points belonging to $v_{j}$.
These offsets are then processed by a point head to generate point-wise features.

\subsection{Beam density estimation module}
\label{sec:BeamEst}
In this section, we detail the method to compute the density expectation for each ray emitted from a LiDAR sensor.
Since all rays from a LiDAR originate from a singular source and are emitted at a fixed angle, the density associated with each beam can be deduced using a spherical projection informed by the beam configuration.
The LiDAR sensor configuration, accessible from low-level sources such as an SDK, determines the set of horizontal and vertical inclinations $\mathbf{C}_h$ and $\mathbf{C}_v$ for each beam in this manner:
\begin{equation}
\begin{split}
    \mathbf{C}_{h} = \left\{ \frac{2\pi i}{H_{\text{b}}} \right\}_{ i \in \{1, ... ,H_{\text{b}}\}},~
    \mathbf{C}_{v}=\left\{\frac{(f_{\text{max}}-f_{\text{min}})j}{V_{\text{b}}}  + f_{\text{min}} \right\}_{j \in \{1,...,V_{\text{b}}\}},
\end{split}
\label{eq:config}
\end{equation}
where $f_{\text{min}}$ and $f_{\text{max}}$ denote the minimum and maximum LiDAR field of view, while $H_{\text{b}}$ and $V_{\text{b}}$ represent the number of horizontal and vertical beams, detailed in~\cref{tab:config}.
To project LiDAR beam inclinations onto spherical projected image coordinates,
we use the projection function $\pi : (\theta, \phi) \to (\dot{\theta}, \dot{\phi})$ defined as:
\begin{equation}
\begin{split}    
     \dot{\theta} =  \left\lfloor \frac{\theta }{2 \pi} W  \right\rfloor, \
     \dot{\phi} = \left\lfloor \frac{\phi - f_{\text{min}}^{\text{proj}}}{f_{\text{max}}^{\text{proj}} - f_{\text{min}}^{\text{proj}}} H \right\rfloor,
    \label{eq:proj}    
\end{split}
\end{equation}
where $H$ and $W$ are the height and width resolutions of the projected image, respectively set to $H=512$, $W=5120$. We define the projected image's field of view as $\left[f_{\text{min}}^{\text{proj}},f_{\text{max}}^{\text{proj}}\right] = \left[-30.0, 15.0\right]$ to accommodate a range of LiDAR sensors.
Given this, beam configurations can be transformed into 1-D binary vectors ${\mathbf{B}}_v \in \mathbbm{R}^{H}$ and ${\mathbf{B}}_h \in \mathbbm{R}^{W}$ as follows:
\begin{equation}
\begin{aligned}
    {\mathbf{B}}_h(\dot{\theta}) =  
    \mathbbm{1}_{\mathbf{C}_{h}}(\theta),~
    {\mathbf{B}}_v(\dot{\phi}) =  
    \mathbbm{1}_{\mathbf{C}_{v}}(\phi), \,\,
    \text{where}~
    \mathbbm{1}_\mathbf{C}(x) = \begin{cases}
    1, & \text{if } x \in \mathbf{C} \\
    0, & \text{otherwise}
  \end{cases},  
\label{eq:density1}
\end{aligned}
\end{equation}
where the indicator function $\mathbbm{1}(\cdot)$ yields 1 when the azimuth or elevation $(\theta, \phi)$ of a pixel $(\dot{\theta}, \dot{\phi})$ corresponds to the beam inclinations specified in $\mathbf{C}$ in~\cref{eq:config}. 
Here, $\mathbf{C}$ can be either $\mathbf{C}_h$ or $\mathbf{C}_v$ that represent the locations of projected vertical and horizontal beams, respectively.
To compute the beam density, we convolve the binary vectors $\mathbf{B}_h$ and $\mathbf{B}_v$, with four distinct 1-D Gaussian kernels, each characterized by standard deviations $\sigma_{k}=\{10, 30, 50, 70\}$ as follows:
\begin{equation}
\begin{aligned}
    {\hat{\mathbf{B}}}_h={\mathbf{B}}_h \ast \mathbf{G}_{\sigma_{k}}, ~{\hat{\mathbf{B}}}_v={\mathbf{B}}_v \ast \mathbf{G}_{\sigma_{k}}.
\end{aligned}
\end{equation}
Finally, we define beam density $\mathcal{D}_i$ of point $p_i$ as follows:
\begin{equation}
\begin{split}
    \mathcal{D}_{i} = \left[ \sqrt{{\hat{\mathbf{B}}}_h^{(k)}(\dot{\theta_i})\cdot{\hat{\mathbf{B}}}_v^{(k)}(\dot{\phi_i})/{r^2_{i}}} \right]_{k=1}^4,
\end{split}
\label{eq:density_main}
\end{equation}
where $r_i$ is the radial distance of a LiDAR point $p_i$ in the spherical coordinates, and $[\cdot]_{k=1}^4$ denotes the concatenation operation.
Taking into account that the density of the beam diminishes in proportion to the square of the distance due to its radial emission, we incorporate the inverse of $r^2$ in our computation.

\subsection{Density soft clipping} 
\label{sec:Density clipping}
When exposed to densities that deviate from those in the source dataset, the model is susceptible to performance degradation in an unseen domain.
To alleviate this issue, we introduce a density soft clipping method that confines the density spectrum using the $\tanh(\cdot)$ function as follows:
\begin{equation}
\begin{split}
   \mathcal{D}_{i}^{c} = \tanh&\left(\frac{\mathcal{D}_{i}-m}{l}\right) l + m,\\
   m=\frac{\mathcal{P}_{90}(\mathcal{D})+\mathcal{P}_{10}(\mathcal{D})}{2},&~l=\frac{\mathcal{P}_{90}(\mathcal{D})-\mathcal{P}_{10}(\mathcal{D})}{2},
    \label{eq:clip}
\end{split}
\end{equation}
where $\mathcal{D}_{i}^{c}$ is clipped density of point $p_i$, and $\mathcal{P}_{90}$ and $\mathcal{P}_{10}$ are functions that extract $90^{\text{th}}$ and $10^{\text{th}}$ channel-wise percentile values from the density embedding of the training domain, respectively.
Employing these percentile values allows us to discount the outlier density values.
To calculate the percentile on the fly under memory constraints, we adopt the Reservoir Sampling~\cite{Vitter1985Reservoir} technique with a sample size $N=1000$.
Please refer to the supplementary material for a more detailed algorithm description.

\subsection{Density-aware embedding module}
\label{sec:DensityEmb}
For each point $p_i$, a point-wise density embedding feature $\mathcal{D}_{i}^{c}$ is obtained via the beam density estimation module incorporating density soft clipping.
This feature serves as an input of both a point-wise attention function $f_{p}$ and a voxel-wise attention function $f_{v}$. 
These mechanisms are pivotal for crafting domain-invariant density-discriminative features.
In the point-wise attention framework, $\mathcal{D}_{i}^{c}$ is synchronized dimensionally with the point-wise feature $F^{p}_i$, facilitating the creation of a density discriminative feature $\hat{F}^{p}_i$ as follows:
\begin{equation}
    \hat{F}^{p}_{i} = f_p(\mathcal{D}_{i}^{c})\odot F^{p}_{i},
\end{equation}
where $f_p$ consists of two 1D convolution layers followed by a sigmoid function.
For voxel-wise attention, it begins by averaging the density embedding features for all points encapsulated within a voxel. This aggregated measure then undergoes a similar treatment as follows:
\begin{equation}
\begin{split}
    \hat{F}^{v}_{j} = \text{Concat}(f_v(\mathcal{D}_{j}^{c})\odot F^{v}_{j}, g(F^{p}_{j})),~\mathcal{D}_{j}^{c} = \frac{1}{|L_{j}|} \sum_{p_i \in v_{j}} \mathcal{D}_{i}^{c}, 
\end{split}
\end{equation}
where $\mathcal{D}_{j}^{c}$ is a clipped density of voxel $v_j$ and $\hat{F}^{v}_{j}$ is a voxel-wise density discriminative feature and $f_v$ consists of two 1D convolution layers followed by a sigmoid function.
An aggregation function $g$ for the voxel-wise feature $F^{v}_{j}$ is adjusted by voxel-wise attention and concatenated with the max pooled point-wise feature within the $v_{j}$. 
Finally, the extracted features are passed through a single 1D convolutional layer, resulting in 32-channel features. 
These density-aware features are subsequently fed into a 3D backbone for LiDAR semantic segmentation.

\subsection{Density Augmentation}
\label{sec:data_augmentation}
The DDFE module aims to align densities across varied sensor domains by considering each dataset as a collection of multiple density domains. 
This method, while effective in many scenarios, may face challenges when there's little to no overlap in the density ranges between the source and unseen datasets. 
To address this potential issue, we introduce a density augmentation method aimed at broadening the density spectrum covered by the training data.
Several existing 3D point cloud augmentation strategies, such as~\cite{Nekrasov2021MIX3D, wei2022beamdrop, Kim2023SDDG} outlined in recent studies, attempt to address density variations either directly or by implication.
We adapt the Mix3D~\cite{Nekrasov2021MIX3D} for our purposes, which we refer to as enhanced-Mix3D, incorporating random translations along the direction of ego-vehicle movement, along with additional rotational transformations to simulate a wider range of density variations.
We also employ beam sampling technique~\cite{wei2022beamdrop, Kim2023SDDG}, selectively eliminating specific LiDAR beams, to amplify the density in the lower direction.
During the training phase, we apply both enhanced-Mix3D and beam sampling augmentations with a set probability of 0.5, aiming to effectively simulate diverse density conditions. Please refer to the supplementary material for more detailed implementation.

\section{Experiments}
In this section, we demonstrate the domain generalization performance of our method through extensive experiments. 
The implementation details and experimental settings of the proposed method are detailed in~\cref{sec:details}.
The configuration of datasets used for evaluation is detailed in~\cref{sec:datasetss}.
A comparative analysis with recent domain generalization and domain adaptation methods is presented in~\cref{sec:comparison}.
The effectiveness of individual components within the proposed DDFE and computational cost are analyzed in Section~\cref{sec:ablation}. 
Detailed per-class performance metrics are available in the supplementary material.

\subsection{Implementation Details}
\label{sec:details}
We employ the point head inspired by Cylinder3D~\cite{Zhu2021Cylindrical} to produce point-wise outputs as shown in \cref{figure_overview}.
During the training phase, we incorporate the Lovasz-Softmax loss $\mathcal{L}^{lovasz}$~\cite{Berman2018Lovasz} along with the Weighted Cross-Entropy loss $\mathcal{L}^{wce}$. The total loss is composed of an equal-weighted combination of point-wise and voxel-wise losses as follows:
\begin{gather}
\begin{split}
    \centering
    \mathcal{L}_{total} = \mathcal{L}_{point}^{lovasz} + \mathcal{L}_{point}^{wce} + \mathcal{L}_{voxel}^{lovasz} + \mathcal{L}_{voxel}^{wce}.
    \label{eq:zeroaug}
\end{split}
\end{gather}
We use the Adam optimizer with an initial learning rate of 1e-3. This rate is decreased by a factor of 0.99 with every epoch.
The training is conducted over 30 epochs with a batch size of two on a single NVIDIA RTX 3090, and the epoch yielding the highest source validation mIoU is chosen. For voxelization, we adopt a cubic size of [20cm, 20cm, 20cm].
In cases where multiple point labels are found within a voxel, the voxel label is determined based on the predominant label, aligning with the approach in Cylinder3D~\cite{Zhu2021Cylindrical}.

\subsection{Datasets}
\label{sec:datasetss}
In our experimental setup, we employ three datasets: Waymo~\cite{Sun2020waymo}, nuScenes~\cite{Caesar2020NUSCENES}, and SemanticKITTI~\cite{Behley2019SemanticKITTI}.
Notably, the Waymo and SemanticKITTI datasets use 64-channel LiDAR, while the nuScenes employs a 32-channel LiDAR.
For detailed information, please refer to~\cref{tab:config}.
In line with methods from previous studies~\cite{Ryu2023LiDomAug, Kim2023SDDG}, we split the nuScenes sequences into 700 for training and 150 for validation.
The Waymo dataset is partitioned into 798 sequences for training and 202 for validation. Regarding the SemanticKITTI dataset, we use sequences 00 to 10 for training, setting aside sequence 08 exclusively for validation.
In integrating class variations across datasets, we follow the class mapping configurations from previous studies for fair comparisons.

\subsection{Comparison to State-of-the-Art DA/DG Methods}
\label{sec:comparison}
The field of domain generalization of LiDAR-based semantic segmentation is still in its formative stages, with a lack of universally accepted experimental standards.
This situation has led to a diversity of experimental designs across different studies. To facilitate fair comparisons, we align our experimental framework with those utilized in seminal works such as LiDomAug~\cite{Ryu2023LiDomAug} and DGLSS~\cite{Kim2023SDDG}.
This approach allows us to benchmark our method against a range of Domain Adaptation (DA) and Domain Generalization (DG) techniques, showcasing the effectiveness of our method in the context of evolving domain generalization challenges.

\begin{table*}[t]
\caption{Comparison with domain generalization methods based on MinkowskiNet architecture using the mIoU. The best and the second best results are highlighted in \textbf{bold} and \underline{underline}, respectively.}
\centering
\resizebox{0.98\textwidth}{!}{%
\begin{tabular}{l|c||c|ccc||c|ccc||c|ccc}
Method     & DA & Source             & W              & K              & N              & Source             & K              & W              & N              & Source             & N              & W              & K              \\ \hline
Base       &    & \multirow{8}{*}{W} & 75.37          & 49.40          & 47.83          & \multirow{8}{*}{K} & 57.31          & 35.24          & 37.42          & \multirow{8}{*}{N} & \underline{65.78} & 38.65          & 36.24          \\
IBN-Net~\cite{pan2018IBN}    &    &                    & \underline{75.47} & 51.13          & 44.72          &                    & 57.74          & 36.99          & 38.74          &                    & 65.31          & 36.53          & 36.93          \\
MLDG~\cite{li2018MLDG}       &    &                    & 72.47          & 48.94          & 48.64          &                    & 56.26          & 35.39          & 36.77          &                    & 61.32          & 36.33          & 32.70          \\
COSMIX (W)~\cite{Saltori2022COSMIX} &  \ding{51}  &                    & -              & -              & -              &                    & 49.35          & 39.46          & 38.94          &                    &  -              & -              & -              \\
COSMIX (K)~\cite{Saltori2022COSMIX} &  \ding{51}  &                    & 66.68          & 44.71          & \underline{49.96} &                    & -              & -              & -              &                    & -               & -              & -              \\
COSMIX (N)~\cite{Saltori2022COSMIX} &  \ding{51}  &                    & 65.68          & 40.99          & 47.98          &                    & 49.98          & 38.05          & 43.25          &                    & -               & -              & -              \\
DGLSS~\cite{Kim2023SDDG}      &    &                    & 75.28          & \underline{51.23} & 49.61          &                    & \underline{59.62} & \underline{40.67} & \underline{44.83} &                    & 65.32          & \underline{40.93} & \underline{38.98} \\
\cellcolor[gray]{0.93}Ours       & \cellcolor[gray]{0.93}   &                    & \cellcolor[gray]{0.93}\textbf{76.15} & \cellcolor[gray]{0.93}\textbf{57.07} & \cellcolor[gray]{0.93}\textbf{56.75} &                    & \cellcolor[gray]{0.93}\textbf{62.50} & \cellcolor[gray]{0.93}\textbf{42.73} & \cellcolor[gray]{0.93}\textbf{49.43} &                    & \cellcolor[gray]{0.93}\textbf{68.16} & \cellcolor[gray]{0.93}\textbf{45.98} & \cellcolor[gray]{0.93}\textbf{46.52}
\end{tabular}
}
\label{tab-dglss}
\end{table*}

\subsubsection{Experiments in the DGLSS Setting}
We compare our method with a domain adaptation method~\cite{Saltori2022COSMIX} and three domain generalization methods~\cite{pan2018IBN, li2018MLDG, Kim2023SDDG}, following the experimental framework defined in DGLSS~\cite{Kim2023SDDG} using Waymo (W), SemanticKITTI (K), and nuScenes (N).
We adopt MinkowskiNet~\cite{Choyi2019MINKOWSKI} as our backbone network, aligning with the configuration utilized in DGLSS.  
The results in~\cref{tab-dglss} indicate that the proposed method consistently outperforms DGLSS across all datasets.
Impressively, our method not only demonstrates superior domain generalization performance but also dominates in the source-to-source settings (W$ \rightarrow $W, K$ \rightarrow $K, N$ \rightarrow $N).
Unlike DGLSS, which tailors its augmentation strategies for each specific dataset, our method employs a uniform hyperparameter setting across all datasets, achieving enhanced domain generalization results. 
Our method achieves an average increase of +12.9\% over DGLSS for unseen datasets using Waymo as the source data (W$ \rightarrow $K, W$ \rightarrow $N).
With SemanticKITTI as the source, there's an average enhancement of +7.8\% for other unseen domains (K$ \rightarrow $W, K$ \rightarrow $N), and employing nuScenes as the source data yields an average increase of +15.8\% on other unseen datasets (N$ \rightarrow $W, N$ \rightarrow $K).

These results demonstrate the robustness of the proposed method in addressing domain discrepancies.

\begin{table}[t]
\caption{Comparison with domain adaptation and data augmentation methods.}
\centering
\resizebox{0.8\textwidth}{!}{%
\begin{tabular}{c|l|c|c|l|c|l|c|c}
Backbone              & \multicolumn{1}{c|}{Methods} & K$ \rightarrow $N & N$ \rightarrow $K &  & Backbone            & \multicolumn{1}{c|}{Methods} & K$ \rightarrow $N & N$ \rightarrow $K \\ \cline{1-4} \cline{6-9} 
\multirow{8}{*}{MinkNet42~\cite{Choyi2019MINKOWSKI}} & Base                         & 37.8 & 36.1 &  & \multirow{8}{*}{C\&L~\cite{Yi2021Complete}} & Base                         & 27.9 & 23.5 \\
                      & CutMix~\cite{Yun2019CutMix}                       & 37.1 & 37.6 &  &                     & SWD~\cite{Lee2019SWD}                          & 27.7 & 24.5 \\
                      & Copy-Paste~\cite{Ghiasi2021COPYPASTE}                   & 38.5 & 41.1 &  &                     & 3DGCA~\cite{Wu2019Squeezesegv2}                        & 27.4 & 23.9 \\
                      & Mix3D~\cite{Nekrasov2021MIX3D}                        & 43.1 & 44.7 &  &                     & C\&L~\cite{Yi2021Complete}                           & 31.6 & 33.7 \\
                      & PolarMix~\cite{Xiao2021POLARMIX}                     & 45.8 & 39.1 &  &                     & LiDomAug~\cite{Ryu2023LiDomAug}                     & 39.2 & 37.9 \\
                      & LiDomAug~\cite{Ryu2023LiDomAug}                     & 45.9 & \underline{48.3} &  &                     & LiDAR-UDA~\cite{shaban2023lidaruda}                    & 41.8 & 34.0 \\
                      & Ours (v=5cm)                         & \underline{48.6} & \textbf{51.3} &  &                     & Ours (v=5cm)                         & \underline{42.5} & \textbf{41.0}\\
                      & \cellcolor[gray]{0.93}Ours (v=20cm)                        & \cellcolor[gray]{0.93}\textbf{50.1} & \cellcolor[gray]{0.93}46.3 &  &                     & \cellcolor[gray]{0.93}Ours (v=20cm)                          & \cellcolor[gray]{0.93}\textbf{47.1} & \cellcolor[gray]{0.93}\underline{40.3}
\end{tabular}
}
\label{table-lidomaug-mink}
\end{table}

\subsubsection{Experiments in the LiDomAug Setting} 
We benchmark the proposed method against various augmentation methods~\cite{Yun2019CutMix, Ghiasi2021COPYPASTE, Nekrasov2021MIX3D, Xiao2021POLARMIX}, domain adaptation methods~\cite{Lee2019SWD, Wu2019Squeezesegv2, Yi2021Complete, shaban2023lidaruda}, and a domain generation method~\cite{Ryu2023LiDomAug}.
We adhere to the experimental setup established by LiDomAug~\cite{Ryu2023LiDomAug}, employing MinkNet42~\cite{Choyi2019MINKOWSKI} and C\&L~\cite{Yi2021Complete} as backbone networks, and utilizing SemanticKITTI and nuScenes datasets for evaluation.
Unlike the DGLSS~\cite{Kim2023SDDG} setting which uses a voxel size of 20 cm, the LiDomAug setting uses a voxel size of 5 cm.
We benchmark the proposed method against various augmentation methods~\cite{Yun2019CutMix, Ghiasi2021COPYPASTE, Nekrasov2021MIX3D, Xiao2021POLARMIX}, domain adaptation methods~\cite{Lee2019SWD, Wu2019Squeezesegv2, Yi2021Complete, shaban2023lidaruda}, and a domain generation method~\cite{Ryu2023LiDomAug}.
We adhere to the experimental setup established by LiDomAug~\cite{Ryu2023LiDomAug}, employing MinkNet42~\cite{Choyi2019MINKOWSKI} and C\&L~\cite{Yi2021Complete} as backbone networks, and utilizing SemanticKITTI and nuScenes datasets for evaluation.
The comparative analysis in \cref{table-lidomaug-mink} shows that our method surpasses all considered augmentation methods in both (K$ \rightarrow $N) and (N$ \rightarrow $K) configurations.
Particularly with MinkNet42 as the backbone, our method achieves a 5.9\% and 5.8\% increase in performance over LiDomAug in the (K$ \rightarrow $N) and (N$ \rightarrow $K) scenarios, respectively.
This enhancement is significant, considering LiDomAug's reliance on ego-motion and multi-frame data integration for domain generalization.
When utilizing C\&L~\cite{Yi2021Complete} as the backbone, the proposed method significantly improves mIoU by +34.5\% and +8.4\% for the scenario of training on SemanticKITTI and evaluating on nuScenes (K$ \rightarrow $N), compared to C\&L and LiDomAug, respectively.
Moreover, in the transition from the sparse 32-channel dataset (nuScenes) to the denser 64-channel dataset (SemanticKITTI) (N$ \rightarrow $K), our method shows a +21.7\% increase in mIoU over C\&L and +8.2\% improvement over LiDomAug.

\newlength{\imageheight}
\setlength{\imageheight}{2.57cm}

\begin{figure}[t]
    \centering
    \begin{minipage}{.02\textwidth}
        \centering
        \rotatebox{90}{N$ \rightarrow $K}
    \end{minipage}
    \begin{minipage}{.96\textwidth}
        \begin{subfigure}{.32\linewidth}
            \centering
            \includegraphics[width=\linewidth, height=\imageheight]{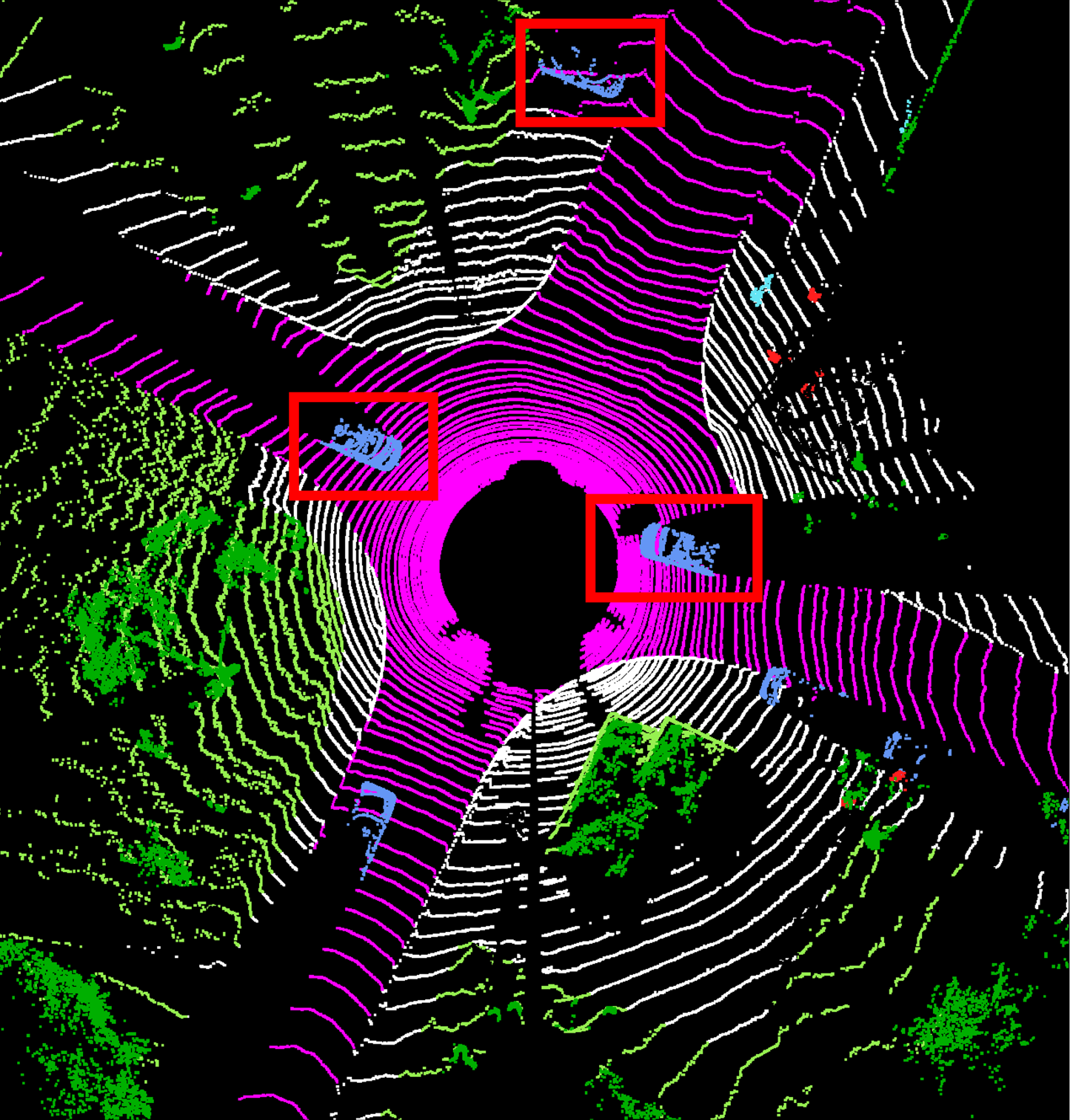}
        \end{subfigure}
        \begin{subfigure}{.32\linewidth}
            \centering
            \includegraphics[width=\linewidth, height=\imageheight]{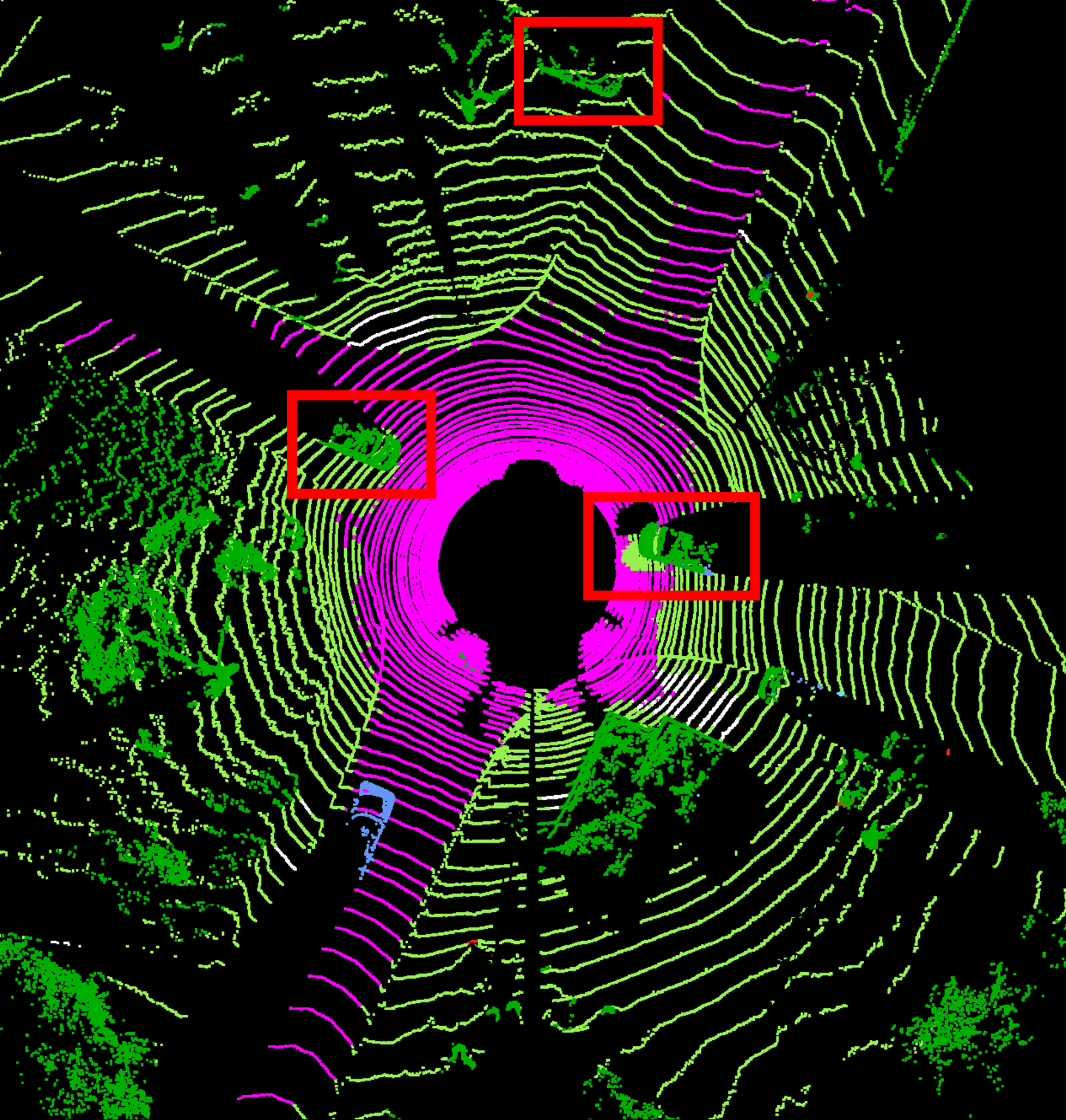}
        \end{subfigure}
        \begin{subfigure}{.32\linewidth}
            \centering
            \includegraphics[width=\linewidth, height=\imageheight]{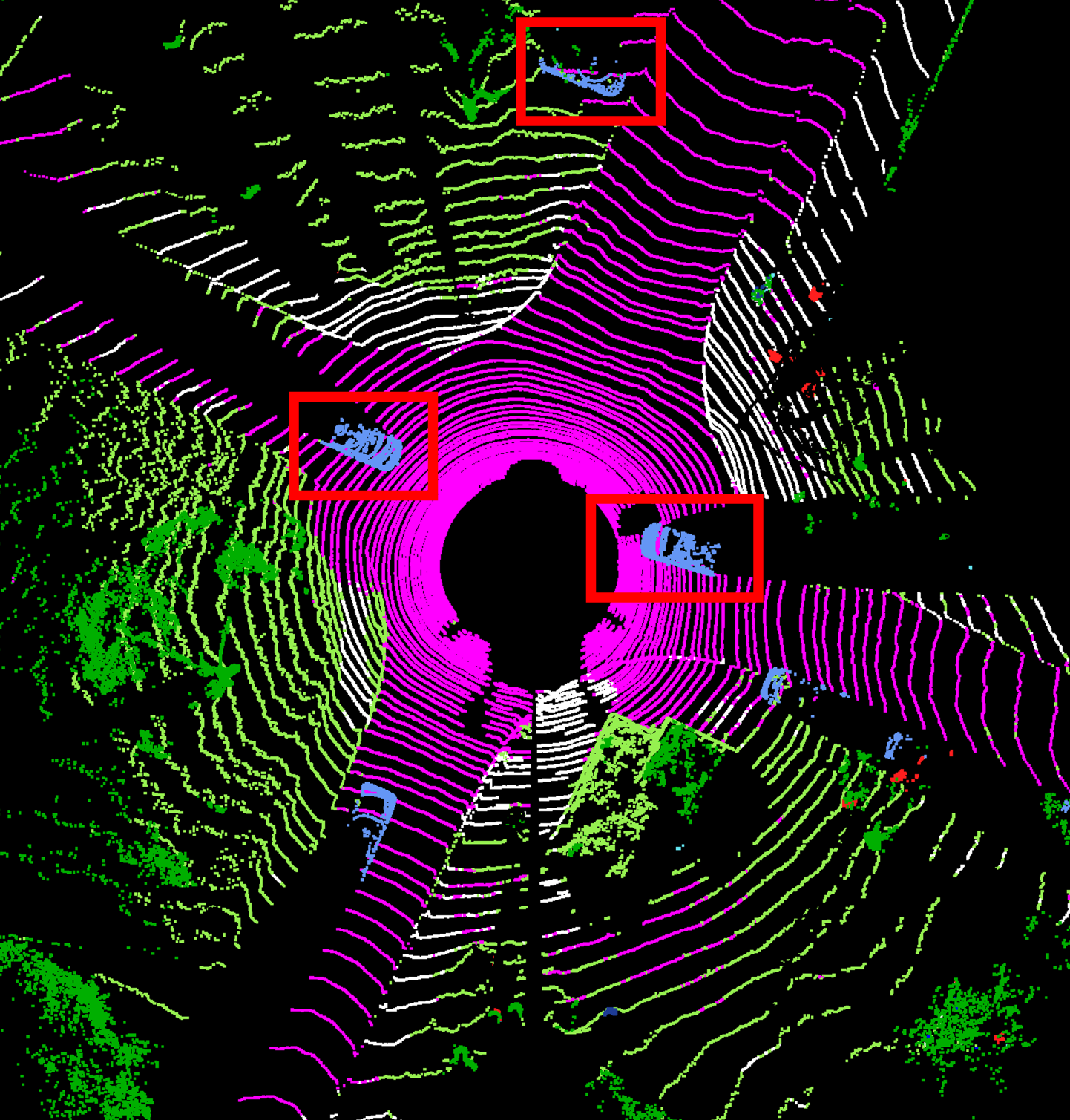}
        \end{subfigure}
    \end{minipage}

    \vspace{0.1cm} 
    
    \begin{minipage}{.02\textwidth}
        \centering
        \rotatebox{90}{K$ \rightarrow $N}
    \end{minipage}
    \begin{minipage}{.96\textwidth}
        \begin{subfigure}{.32\linewidth}
            \centering
            \includegraphics[width=\linewidth, height=\imageheight]{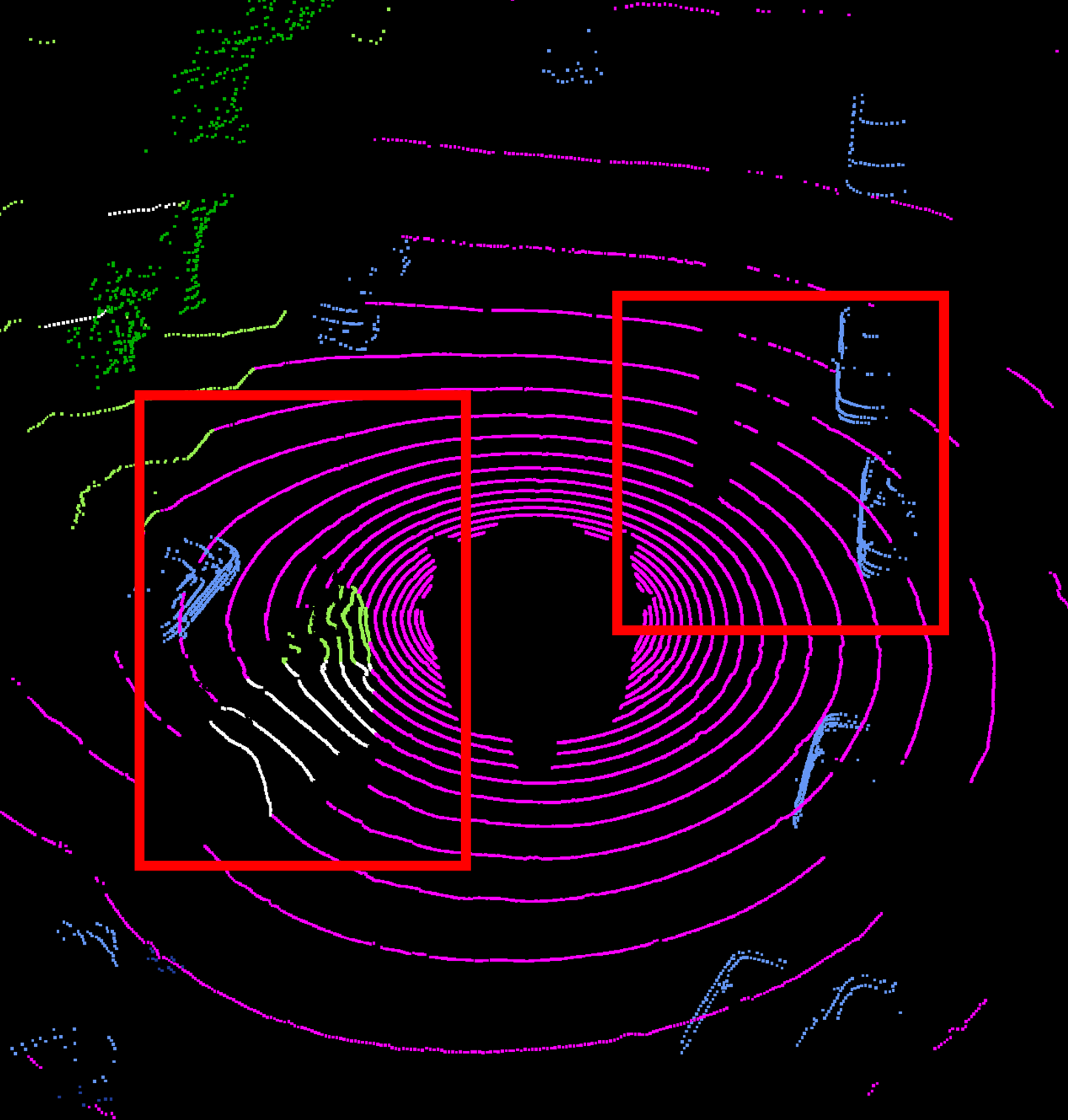}
            \caption*{(a) Ground-truth}
        \end{subfigure}
        \begin{subfigure}{.32\linewidth}
            \centering
            \includegraphics[width=\linewidth, height=\imageheight]{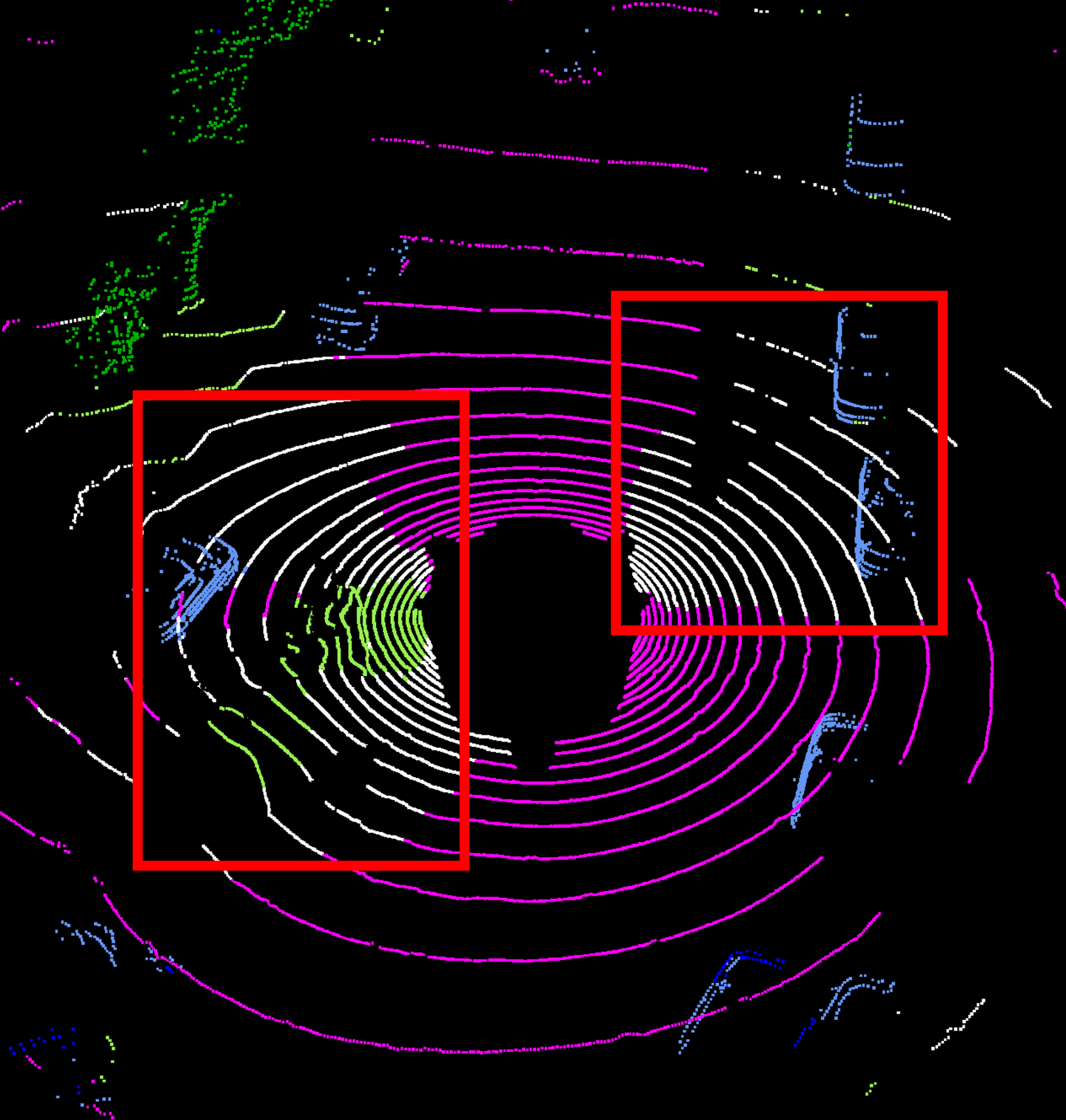}
            \caption*{(b) MinkNet42}
        \end{subfigure}
        \begin{subfigure}{.32\linewidth}
            \centering
            \includegraphics[width=\linewidth, height=\imageheight]{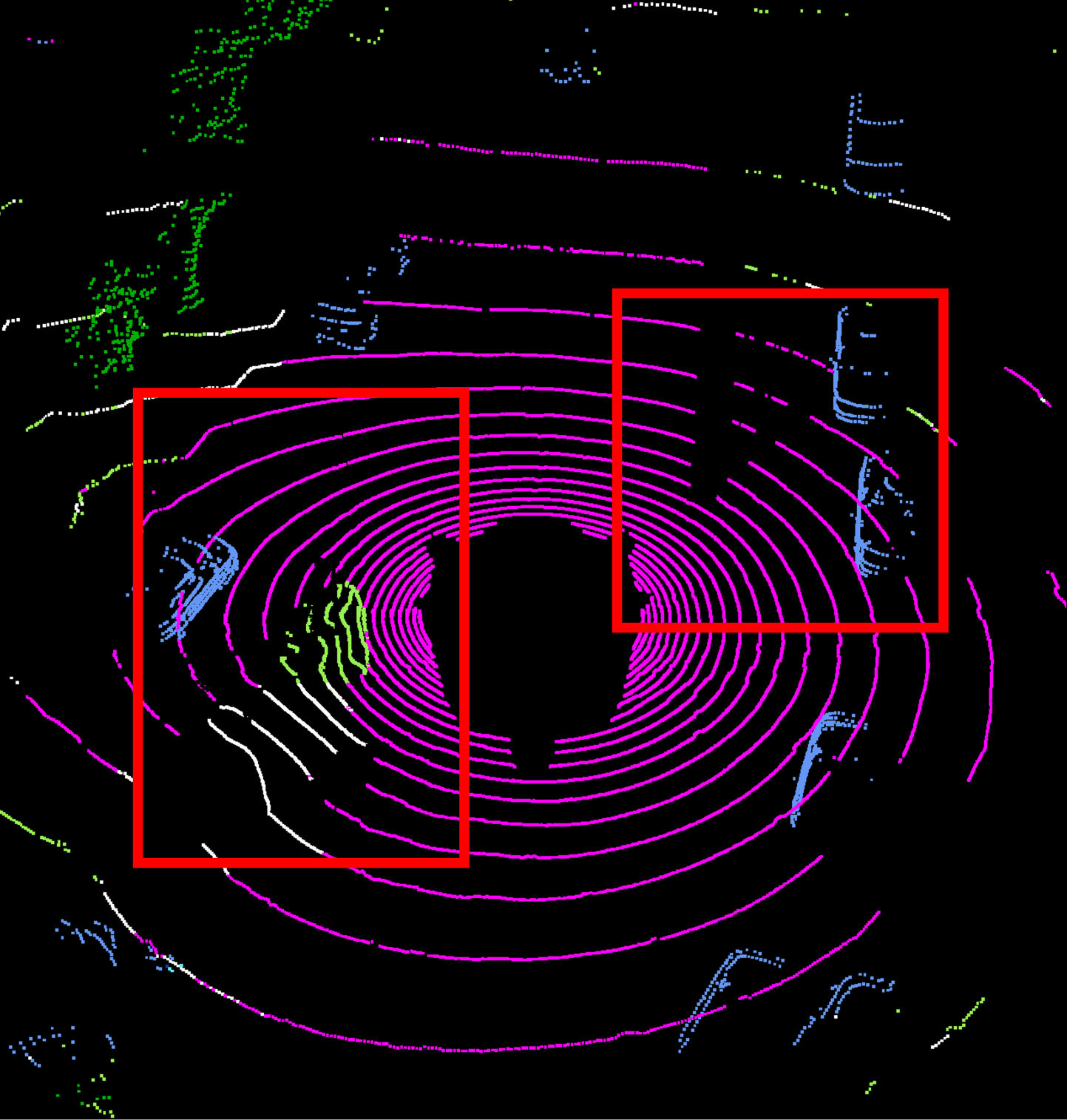}
            \caption*{(c) MinkNet42+Ours}
        \end{subfigure}
    \end{minipage}
\caption{Qualitative comparison with MinkNet42 backbone. Top: Model trained on nuScenes, tested on SemanticKITTI (N$ \rightarrow $K). Bottom: Trained on SemanticKITTI, tested on nuScenes (K$ \rightarrow $N).}
\label{figure_mink_nk}
\end{figure}

We further conduct an analysis with a voxel size of 20 cm, unlike the comparative methods that all use a voxel size of 5 cm.
Increasing the voxel size from 5 cm to 20 cm results in reductions of 30.3\% in training time and 62.5\%  in inference time, while maintaining comparable performance—with a slight decrease of 3.5\% on the MinkNet42 backbone and an increase of 4.7\% on the C\&L backbone. 
Based on these results, we recommend a voxel size of 20 cm (v=20cm) as the default setting for our proposed method.
The effectiveness is further illustrated in the qualitative results depicted in~\cref{figure_mink_nk}, highlighting the substantial improvements our method brings to the domain generalization performance of the 3D backbone network.
Please refer to the supplementary material for detailed results, including per-class IoU.

\begin{table*}[t]
\caption{Ablation study on individual components within our method using mIoU. Experiments evaluate the model's performance with and without the following: (a) Point-voxel encoding, (b) Density-aware embedding module, (c) Density clipping, and (d) Density-Augmentation. All experiments were conducted using a voxel size of 20 cm.}
\centering
{
\begin{tabular}{c|c|c|c|l|l}
(a) & (b) & (c) & (d) & \multicolumn{1}{c|}{K$ \rightarrow $N } & \multicolumn{1}{c}{N$ \rightarrow $K} \\ \hline
\rule{0in}{2ex}  &   &   &       & 40.7              & 31.4              \\
\ding{51}  &   &   &       & 43.0 \textcolor{mygreen}{(+5.7\%)}              & 35.0 \textcolor{mygreen}{(+11.5\%)}              \\
\ding{51}  & \ding{51}  &  &       & 45.7 \textcolor{mygreen}{(+12.3\%)}             & 40.5 \textcolor{mygreen}{(+29.0\%)}             \\
\ding{51}  & \ding{51}  & \ding{51}  &       & 46.2 \textcolor{mygreen}{(+13.5\%)}            & 41.8 \textcolor{mygreen}{(+33.1\%)}              \\
\ding{51}  & \ding{51}  & \ding{51}    & \ding{51}    & \textbf{50.1} \textcolor{mygreen}{(+23.1\%)}              & \textbf{46.3} \textcolor{mygreen}{(+47.5\%)}              \\
\end{tabular}
}
\label{tab:ablation}
\end{table*}

\subsection{Ablation Studies} 
\label{sec:ablation}
To validate the impact of individual components within our method, we conduct ablation studies. These studies concentrate on the evaluation of point-voxel encoding, the density-aware embedding module, density soft clipping, and density augmentation, providing insights into the significance of each element in enhancing the overall methodology.

\subsubsection{Analysis of the individual components of our method}
In our comprehensive ablation study in \cref{tab:ablation}, we thoroughly investigate the individual components of our method. We use the experimental setup proposed by LiDomAug~\cite{Ryu2023LiDomAug} and employ MinkowskiNet~\cite{Choyi2019MINKOWSKI} as the backbone. This study dissects the impact of our DDFE, which includes (a) point-voxel encoding, (b) density-aware embedding module, and (c) density clipping. Additionally, it examines (d) the impact of our density augmentation method.

The integration of all these elements results in a significant performance uplift +13.5\% in the (K$\rightarrow$N) scenario and +33.1\% in the (N$\rightarrow$K) scenario over the baseline model. Notably, the density-aware embedding module stands out for its substantial effect, closely followed by the point-voxel encoding, highlighting its critical role. Furthermore, the study validates the utility of density soft clipping in enhancing model flexibility towards novel density configurations. Meanwhile, the gains observed with density clipping highlight the inherent challenge in adapting to unfamiliar density landscapes, emphasizing the nuanced contributions of each component towards achieving robust domain generalization.
Lastly, incorporating density augmentation into DDFE leads to further enhancements, with an additional 8.4\% performance increase in the (K$\rightarrow$N) scenario and 10.8\% in the (N$\rightarrow$K) scenario. This demonstrates the effectiveness of the proposed density augmentation scheme in significantly boosting the performance of DDFE.

\subsubsection{Augmentation}
\cref{tab-aug} shows the performance comparison when various augmentation methods are applied to the proposed DDFE.
The experimental setup follows DGLSS~\cite{Kim2023SDDG}, and the baseline involves applying the proposed DDFE to MinkowskiNet~\cite{Choyi2019MINKOWSKI}.
The proposed enhanced-Mix3D shows a 2.3\% performance improvement over the original Mix3D~\cite{Nekrasov2021MIX3D} when trained on SemanticKITTI and tested on nuScenes (K$ \rightarrow $N), and a 3.6\% improvement when trained on nuScenes and tested on SemanticKITTI (N$ \rightarrow $K). 
In conjunction with the beam sampling method, it achieves a 7.6\% performance improvement in (K$ \rightarrow $N) and 11.2\% in (N$ \rightarrow $K) compared to the baseline.

\begin{table}[t]
\caption{Comparison of data augmentation methods in the proposed DDFE. The best and the second best results are highlighted in \textbf{bold} and \underline{underline}, respectively. `D. V.' refers to Density Variation situation, and `B. S.' to Beam Sampling augmentation.}
\centering
{
\begin{tabular}{l|c|l|l}
Method  & D. V. & \multicolumn{1}{c|}{K$ \rightarrow $N } & \multicolumn{1}{c}{N$ \rightarrow $K } \\ \hline
DDFE   &    & 45.9          & 41.8                       \\ \hline
+ PolarMix~\cite{Xiao2021POLARMIX}  &   & 48.2 \textcolor{mygreen}{(+5.0\%)}      & 42.3 \textcolor{mygreen}{(+1.2\%)}                    \\
+ B. S.~\cite{wei2022beamdrop}  & \ding{51}      & \underline{49.1} \textcolor{mygreen}{(+7.0\%)}      & 42.2 \textcolor{mygreen}{(+0.9\%)}                      \\
+ Mix3D~\cite{Nekrasov2021MIX3D}  & \ding{51}     & 47.4 \textcolor{mygreen}{(+3.2\%)}       & 44.8 \textcolor{mygreen}{(+7.1\%)}                      \\
+ Mix3D + B. S.~\cite{wei2022beamdrop}  & \ding{51}     & \underline{49.1} \textcolor{mygreen}{(+7.0\%)}       & 44.2 \textcolor{mygreen}{(+5.7\%)}                      \\
+ E-Mix3D  & \ding{51}  & 48.5 \textcolor{mygreen}{(+5.6\%)}      & \underline{46.4} \textcolor{mygreen}{(+11.0\%)}                      \\
+ E-Mix3D + B. S.~\cite{wei2022beamdrop}  & \ding{51}      & \textbf{49.4} \textcolor{mygreen}{(+7.6\%)} & \textbf{46.5} \textcolor{mygreen}{(+11.2\%)} \\
\end{tabular}%
}
\label{tab-aug}
\end{table}

\begin{figure}[!t] 
    \centering
    \includegraphics[width=0.9\columnwidth]{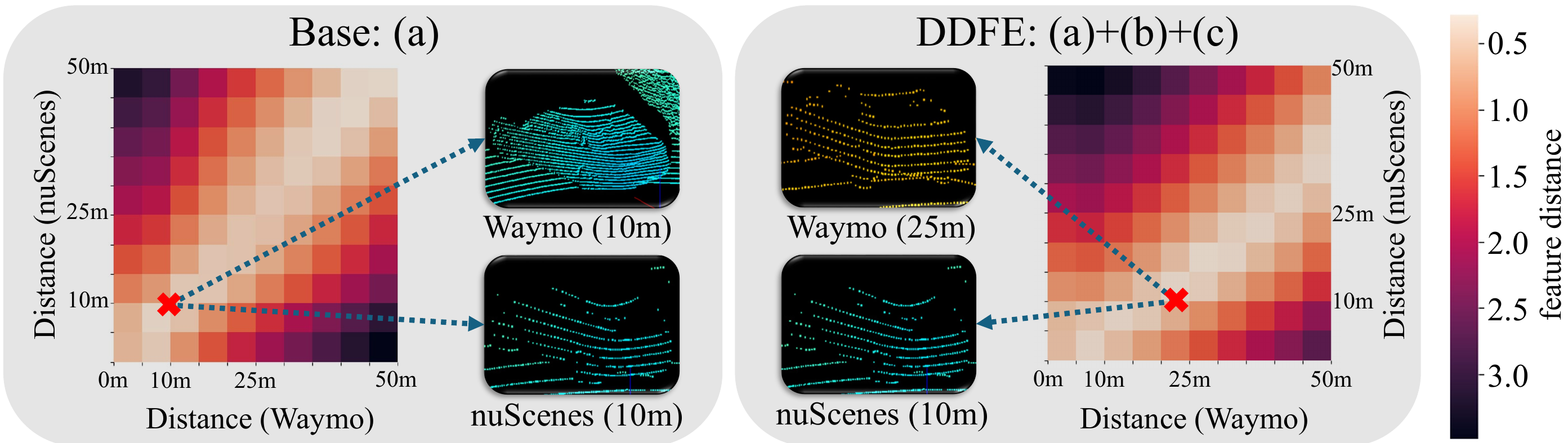}
    \caption{Visualization of feature similarity matrices between the Waymo (64-channel) and the nuScenes (32-channel) datasets, with a focus on the distance of objects from the LiDAR. 
    We utilize models trained with the nuScenes dataset for visualization. (a) A point-voxel feature encoding method. (b) A density-aware embedding module. (c) A density soft clipping. Our DDFE combines all modules (a), (b), and (c).
    }
\label{fig_visualization}
\end{figure}

\subsubsection{Analysis of Density Discriminative Feature Embedding (DDFE)}
We delve into the effectiveness of the DDFE module by examining the feature similarity between the source dataset, nuScenes, and the unseen dataset, Waymo, as depicted in \cref{fig_visualization}.
Our focus is on the input voxel features $\hat{F}^{v}$ of the 3D backbone model generated by the DDFE.
We initiate our analysis by setting a baseline that utilizes a point-voxel feature encoding method for comparison. Following this, we proceed to assess our enhanced method, which integrates a density-aware embedding module along with density soft clipping, to understand its impact on bridging the domain gap.
Similarity metrics are derived from the L2 distance between averaged features across specified distance intervals (\textit{e.g.}, 0-5m, ..., 45-50m) for each dataset. According to \cref{fig_visualization}, the baseline consistently shows higher feature similarity across distances, influenced by its direct embedding of 3D point coordinates. Conversely, our integrated model with density-aware embedding and soft clipping significantly aligns feature similarity across consistent density distributions. It demonstrates the efficacy of the proposed modules in adapting to differences between the source and unseen datasets.

\subsubsection{Computational Cost} 
The inference time of our method for processing a single frame with the nuScenes is 44ms on an NVIDIA RTX 3090. The inclusion of our DDFE module into the MinkNet42 architecture accounts for an extra 8ms of this computation time, while the base MinkNet42 architecture alone requires 36ms. 
This integration modestly elevates the model's complexity, introducing approximately 23.8k parameters (+0.06\%).
Furthermore, the application of density augmentation during training adds an extra 60ms per scan on the nuScenes.

\section{Conclusion}
In this paper, we introduce a new perspective for domain generalization of LiDAR semantic segmentation by exploiting the concept of density diversity within a source domain.
Based on this perspective, we propose the Density Discriminative Feature Embedding (DDFE) module. DDFE is designed to incorporate expected densities derived from LiDAR beams into a density-aware feature space, significantly enhancing the model's capability to distinguish between different densities. This novel approach improves the adaptability and accuracy of LiDAR semantic segmentation, enabling them to perform effectively across diverse domains, including those previously unseen.
In addition to the core methodology, we also present a simple and effective data augmentation technique that extends the density spectrum of the source data.
Extensive experiments on the SemanticKITTI, Waymo, and nuScenes datasets demonstrate the effectiveness of the proposed method in elevating the domain generalization performance of LiDAR semantic segmentation.
A limitation of this study is its exclusive focus on semantic segmentation tasks.
Thus, in future work, we plan to extend our domain generalization method to encompass other LiDAR-based perception tasks, such as 3D object detection.
This extension can involve utilizing object-centric density discriminative features alongside point-wise density embedding.

\section*{Acknowledgements}
This work was supported by the Digital Innovation Hub project supervised by the Daegu Digital Innovation Promotion Agency(DIP) grant funded by the Korea government(MSIT and Daegu Metropolitan City) in 2023(DBSD1-02, Vision picking system for the logistics industry based on artificial intelligence object recognition) and the National Research Foundation of Korea (NRF) grant funded by the Korea government (MSIT) (No. RS-2023-00210908).

\bibliographystyle{splncs04}
\bibliography{main}

\end{document}